\documentclass[]{spie}  

\usepackage{amsmath,amsfonts,amssymb}
\usepackage{graphicx}
\usepackage{subcaption}
\usepackage{multirow}
\usepackage[colorlinks=true, allcolors=blue]{hyperref}

\title{Benchmarking Suite for Synthetic Aperture Radar Imagery Anomaly Detection (SARIAD) Algorithms}

\author[a,b]{Lucian Chauvin}
\author[c]{Somil Gupta}
\author[c]{Angelina Ibarra}
\author[c]{Joshua Peeples}
\affil[a]{Department of Computer Science, Texas A\&M University, College Station, TX, 77845}
\affil[b]{Department of Mathematics, Texas A\&M University, College Station, TX, 77845}
\affil[c]{Department of Electrical and Computer Engineering, Texas A\&M University, College Station, TX, 77845}

\authorinfo{Further author information: (Send correspondence to J.P.)\\J.P.: E-mail: jpeeples@tamu.edu, Telephone: 1 979 458 7026}

\begin{document} 
\maketitle

\begin{abstract}
Anomaly detection is a key research challenge in computer vision and machine learning with applications in many fields from quality control to radar imaging. In radar imaging, specifically synthetic aperture radar (SAR), anomaly detection can be used for the classification, detection, and segmentation of objects of interest. However, there is no method for developing and benchmarking these methods on SAR imagery. To address this issue, we introduce SAR imagery anomaly detection (SARIAD). In conjunction with Anomalib, a deep-learning library for anomaly detection, SARIAD provides a comprehensive suite of algorithms and datasets for assessing and developing anomaly detection approaches on SAR imagery. SARIAD specifically integrates multiple SAR datasets along with tools to effectively apply various anomaly detection algorithms to SAR imagery.  Several anomaly detection metrics and visualizations are available. Overall, SARIAD acts as a central package for benchmarking SAR models and datasets to allow for reproducible research in the field of anomaly detection in SAR imagery. This package is publicly available: \url{https://github.com/Advanced-Vision-and-Learning-Lab/SARIAD}.
\end{abstract}



\section{INTRODUCTION}
\label{sec:intro}  
Synthetic aperture radar (SAR) is a powerful remote sensing technology that provides high-resolution imaging capabilities regardless of weather conditions or lighting \cite{moreira2013tutorial}. Unlike optical sensors, SAR systems actively emit microwave signals and analyze their reflections, making them invaluable for a wide range of applications, including environmental monitoring \cite{amitrano2021earth}, surveillance \cite{garcia2024advancements}, and disaster response \cite{dunkel2011synthetic}. Despite these advantages, interpreting SAR imagery presents significant challenges due to noise, speckle effects, and complex scene variations \cite{javali2021review}.

Anomaly detection in SAR imagery is crucial for identifying unusual patterns or objects of interest, such as detecting concealed targets\cite{palm2021robust}, assessing infrastructure damage \cite{yang2024large}, or monitoring environmental changes \cite{swaroop2024land}. However, SAR anomaly detection remains a difficult problem due to the inherent complexity of SAR data, the lack of large labeled datasets, and the need for robust algorithms that generalize well across different environments. Many existing methods struggle with these challenges, making reproducible research and standardized benchmarking critical to advancing the field.

To address these issues, we introduce SAR Imagery Anomaly Detection (SARIAD), a benchmarking suite designed to facilitate reproducible research in SAR anomaly detection. Our contributions include: 1) a modular framework for the evaluation of SAR datasets, preprocessing, and anomaly detection models, and  2) an approach to generate ``normal" data for datasets that do not have this information readily available.
By providing a comprehensive and standardized framework, SARIAD enables researchers to systematically evaluate SAR anomaly detection methods, fostering innovation and accelerating progress in the field.

\section{RELATED WORK}
\label{sec:rel}  

Anomalib \cite{akcay2022anomalib} is an open-source deep-learning library designed for unsupervised anomaly detection. With a focus on modularity and reproducibility, Anomalib provides state-of-the-art anomaly detection algorithms that can be used off-the-shelf or customized through a plug-and-play approach. It also includes various tools for designing, training, and evaluating anomaly detection models, supporting inference through OpenVINO \cite{gorbachev2019openvino}.
Anomalib addresses the challenge of benchmarking anomaly detection methods by unifying datasets, pre-processing, models, and more within a single framework. Unlike existing anomaly detection libraries that often focus on specific techniques, Anomalib integrates multiple deep learning methods at a high level, making it a comprehensive solution for reproducible research.

Inspired by Anomalib, Benchmarks for Medical Anomaly Detection (BMAD) \cite{bao2024bmad} was introduced for anomaly detection across different biomedical image modalities. In our work, we build upon Anomalib and BMAD by incorporating SAR imagery datasets and methods into the proposed framework, adapting anomaly detection models for use in SAR imagery. Our approach extends Anomalib’s benchmarking capabilities to SAR data, introducing additional pre-processing techniques, datasets, and domain-specific evaluation metrics. Another work related to this work is a package developed for SAR target detection and recognition: GrokSAR \cite{dai2024denodet}. Key differences in our work are that our package aims to be more modular that allows for the incorporation of different models and datasets, better experiment control by allowing for configurable training procedures, and GrokSAR primarily focuses on object detection, while our package evaluates anomaly detection for image- and pixel-level tasks.

\section{METHODS}
\label{sec:methods}  

\subsection{SARIAD Overall Design}
\label{sec:SARIAD}
\begin{figure}[htb]
    \centering
    \begin{subfigure}{0.3\textwidth}
        \centering
        \includegraphics[width=\linewidth]{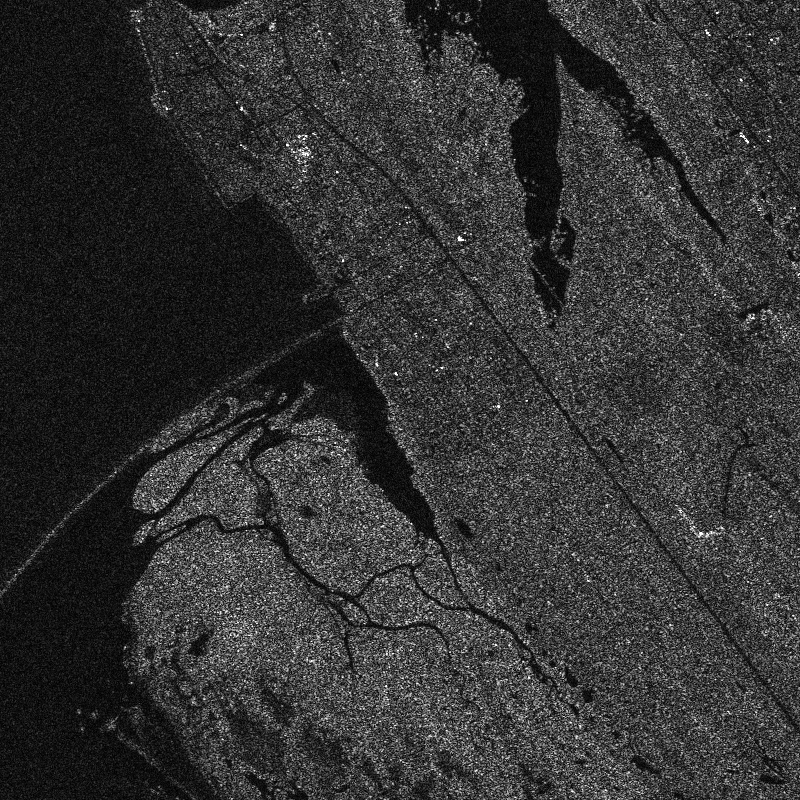}
    \end{subfigure}
    \hfill
    \begin{subfigure}{0.3\textwidth}
        \centering
        \includegraphics[width=\linewidth]{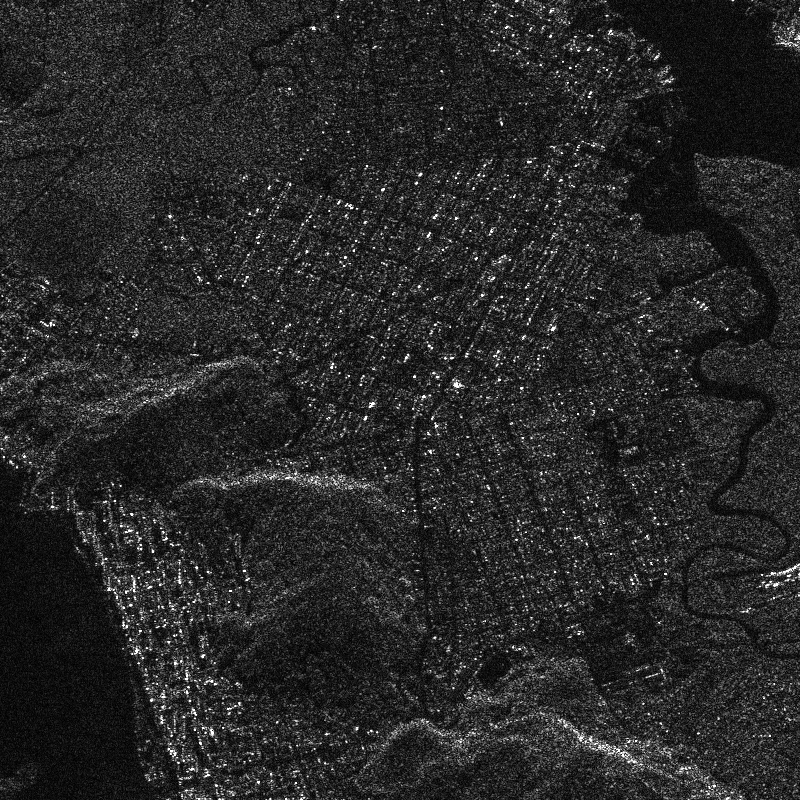}
    \end{subfigure}
    \hfill
    \begin{subfigure}{0.3\textwidth}
        \centering
        \includegraphics[width=\linewidth]{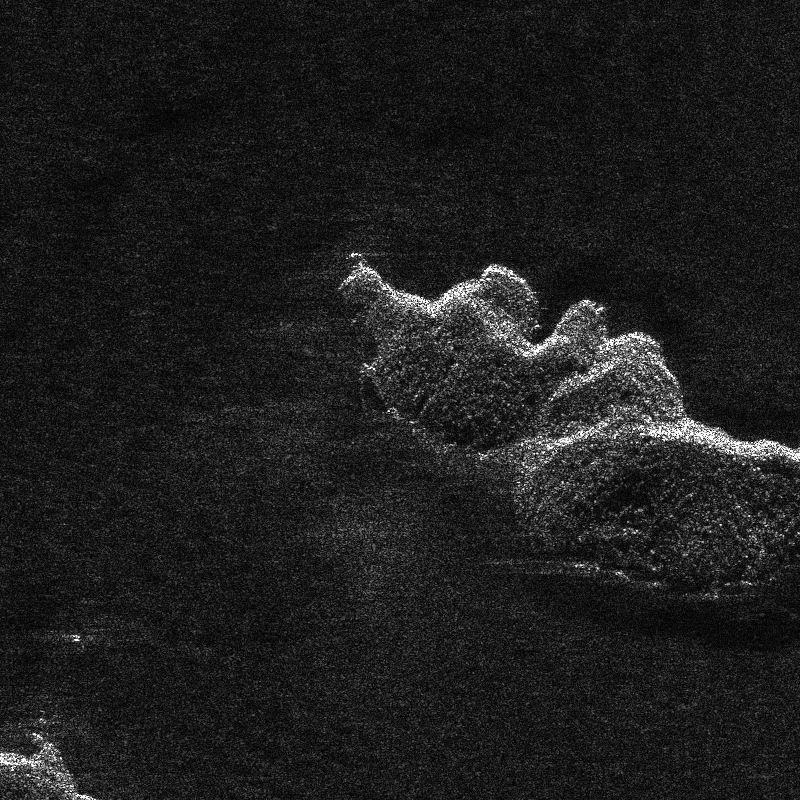}
    \end{subfigure}
    
    \caption{Example images of pure background from HRSID that are used as ``normal" data.}
    \label{fig:HRSID}
\end{figure}

SARIAD overall design follows from Anomalib \cite{akcay2022anomalib}, with each component tailored for SAR. The first aspect of the package is the data module. The data module can be customized for any SAR dataset. The three tasks currently supported by the data module are classification, object detection, and segmentation. In order to use object detection and segmentation, a ground truth must be provided in order to assess each method quantitatively and qualitatively. Additionally, similar to Anomalib, a dataset must have a) normal images only or b) anomalous and normal images for training and evaluation. We implemented two SAR datasets in the initial work: moving and stationary target recognition (MSTAR) \cite{keydel1996mstar} and High-Resolution SAR Images Dataset (HRSID)\cite{wei2020hrsid}. HRSID has 400 images of pure background that can be used as ``normal" data as shown in Figure \ref{fig:HRSID}. The details of the generation of normal data for MSTAR are discussed in Section \ref{sec:normal_data}. Our data modules automatically download these datasets for ease of use.

After the data module, pre-processing can be applied to the images. In this initial work, the images were preprocessed to be in a range of 0 to 1 before being passed into the model. For SAR images, despeckling techniques \cite{di2013benchmarking,vitale2020multi,molini2021speckle2void} can be integrated into our framework to improve the performance of anomaly detection models. Once the datasets are loaded and preprocessed, the model module selects a machine learning algorithm for training, validation, and test. Any model that is available in Anomalib can be used in SARIAD as well as SAR-specific models. Once training and evaluation is completed, a post-processing module produces metrics and visualizations to further assess performance. Finally, the model(s) can be deployed using OpenVINO.

\subsection{MSTAR Normal Data Generation}
\label{sec:normal_data}
\begin{figure}[htb]
    \centering
    \begin{subfigure}{0.15\textwidth}
        \centering
        \includegraphics[width=\linewidth]{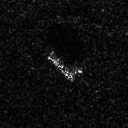}
    \end{subfigure}
    \begin{subfigure}{0.15\textwidth}
        \centering
        \includegraphics[width=\linewidth]{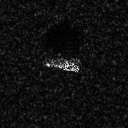}
    \end{subfigure}
    \begin{subfigure}{0.15\textwidth}
        \centering
        \includegraphics[width=\linewidth]{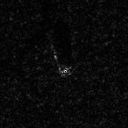}
    \end{subfigure}
    \begin{subfigure}{0.15\textwidth}
        \centering
        \includegraphics[width=\linewidth]{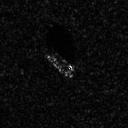}
    \end{subfigure}
    \begin{subfigure}{0.15\textwidth}
        \centering
        \includegraphics[width=\linewidth]{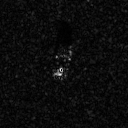}
    \end{subfigure}
    
    \begin{subfigure}{0.15\textwidth}
        \centering
        \includegraphics[width=\linewidth]{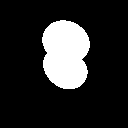}
    \end{subfigure}
    \begin{subfigure}{0.15\textwidth}
        \centering
        \includegraphics[width=\linewidth]{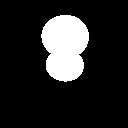}
    \end{subfigure}
    \begin{subfigure}{0.15\textwidth}
    \centering
    \includegraphics[width=\linewidth]{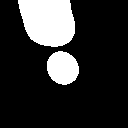}
    \end{subfigure}
    \begin{subfigure}{0.15\textwidth}
    \centering
    \includegraphics[width=\linewidth]{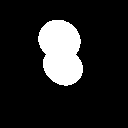}
    \end{subfigure}
    \begin{subfigure}{0.15\textwidth}
    \centering
    \includegraphics[width=\linewidth]{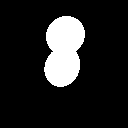}
    \end{subfigure}
    
    \begin{subfigure}{0.15\textwidth}
        \centering
        \includegraphics[width=\linewidth]{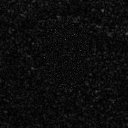}
        \caption{2S1}
    \end{subfigure}
    \begin{subfigure}{0.15\textwidth}
        \centering
        \includegraphics[width=\linewidth]{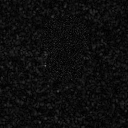}
        \caption{BMP2}
    \end{subfigure}
    \begin{subfigure}{0.15\textwidth}
        \centering
        \includegraphics[width=\linewidth]{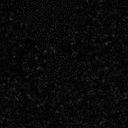}
        \caption{BRDM2}
    \end{subfigure}
    \begin{subfigure}{0.15\textwidth}
        \centering
        \includegraphics[width=\linewidth]{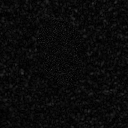}
        \caption{BTR60}
    \end{subfigure}
    \begin{subfigure}{0.15\textwidth}
        \centering
        \includegraphics[width=\linewidth]{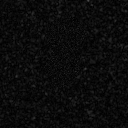}
        \caption{BTR70}
    \end{subfigure}
    
    \caption{Example images of generating ``normal" data for 5 of 10 classes in the MSTAR dataset. MSTAR only has target chips and does not have background images that are needed for SARIAD. The first row is the input images of the target chip. The second row shows the segmentation results of identifying the targets. The last row displays the ``normal" images generated by sampling from the background pixels to fill in the targets.}
    \label{fig:Normal_data}
\end{figure}

To generate the data necessary for training Anomalib models on MSTAR, a systematic approach was used to create masks and normal data. The process started with the SAR image training data set, all of which were anomalous. The first step involved generating masks that highlight anomalous regions of interest. To do this, we used k-Means to group the pixel intensities into two clusters (foreground and background). Once the foreground (\textit{i.e.}, target and shadow) was identified, these pixels were set to 0. The mean and standard deviation of the background pixels were computed to define a Gaussian distribution for background pixels. The area where the target was located is then filled with values sampled from the background distribution. This process is repeated to mask out the shadow of the anomaly. Example images of normal data produced by this method are shown in Figure \ref{fig:Normal_data}. This approach did not always result in consistent localization of the target as the main focus was to ensure that the target and shadow was removed from the normal image. This approach would need to be refined if one would want to evaluate MSTAR for precise pixel-level (\textit{i.e.}, segmentation) or object detection tasks. These methods could be refined by using more iterations when applying k-Means.


\section{EXPERIMENTAL RESULTS and DISCUSSION}
\label{sec:exp}  

To demonstrate the results of our package, we used two datasets, MSTAR \cite{keydel1996mstar} and HRSID \cite{wei2020hrsid}, as mentioned in Section \ref{sec:SARIAD}. MSTAR was used for image-level anomaly detection (\textit{i.e.}, is there a target present or not) while HRSID was evaluated for image-level and pixel-level anomaly (\textit{i.e.}, localization of anomaly). As discussed in Section \ref{sec:normal_data}, MSTAR used k-Means to generate normal data. For obtaining the mask for the anomaly we use k-Means with 2 clusters, 5 different random initializations, and 100 iterations to converge. When we reapply our method to detect the shadow, we invert the image and use k-Means with 5 clusters, 5 different random initializations, and 100 iterations to converge. To capture the shadow of the target, the largest cluster is used to identify the pixels belonging to the shadow of the target. The MSTAR dataset has two operating conditions: standard and extended. The results in this work focus on the standard operating conditions (SOC), but the extended operating conditions (EOC) are available in SARIAD as well. The SOC has a total of 10 targets, while HRSID has several ships of varying sizes contained within the images. Each dataset has predefined partitions for training and testing data. 

For the anomaly detection models, we selected three models that did not require training so we can evaluate the ``zero-shot capability" of these models for SAR: Patch Distribution Modeling (PaDiM)\cite{defard2021padim}, Deep Feature Modeling (DFM) \cite{ahuja2019probabilistic}, and Window-based Contrastive Language–Image Pre-training (WinCLIP) \cite{jeong2023winclip}. The default hyperparameter settings were used for each model. DFM and WinCLIP do not have any randomness associated with each method, but PaDiM randomly selects 100 features from the feature backbone (default is ResNet18). Therefore, a total of five experimental runs were completed for both datasets.

\begin{figure}[htb]
    \centering
    \begin{subfigure}{0.3\textwidth}
        \centering
        \includegraphics[width=\linewidth]{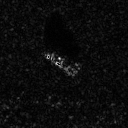}
        \caption{Input S21 Target Image}
    \end{subfigure}
    \begin{subfigure}{0.3\textwidth}
        \centering
        \includegraphics[width=\linewidth]{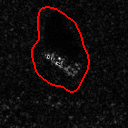}
        \caption{Anomaly Detection Output}
    \end{subfigure}
    
    \caption{Example output resulting from applying PaDiM to a target chip in MSTAR. Despite not having pixel-level ground truth, the model predicts a region of the image where a possible anomaly or target is present. }
    \label{fig:MSTAR_Example}
\end{figure}

The quantitative and qualitative results for MSTAR are shown in Table \ref{tab:MSTAR} and Figure \ref{fig:MSTAR_Example}, respectively. For the three models, WinCLIP performs the best in all metrics except precision when compared to DFM. WinCLIP leverages learned features at multiple scales and this can potentially be attributed to the better performance in comparison to the other models as multi-scaled features have been shown to be effective for anomaly detection \cite{zhou2024msflow}. The PaDiM model also performs well on the MSTAR dataset. PaDiM uses the Mahalanobis distance to assess whether the data is anomalistic. This measure appears to be robust despite the assumption that the data is Gaussian distributed. The MSTAR dataset did not have precise, labeled ground truth; however, as shown in Figure \ref{fig:MSTAR_Example}, PaDiM does a good job at identifying the area of the image where the target is located. This would be useful for an operator in practice to explain why an image is predicted as anomalistic. SARIAD provides a modular package that can be used to evaluate the model across different image metrics and generate clear visuals for explainability. 

\begin{table}[htb] 
\centering
\caption{Average anomaly detection performance metrics on the image-level for the MSTAR dataset. All three models did not require training. DFM and WinCLIP are deterministic in their estimation of anomalistic data. PaDiM randomly selects 100 features from the backbone and this model was used for three experimental trials, resulting in the standard deviation being shown for this model. The best average metric is bolded.} 
\begin{tabular}{|c|c|c|c|c|c|c|} \hline Model & Accuracy (\%) & Precision (\%) & Recall (\%) & F1-Score (\%) & ROC AUC (\%) & PR AUC (\%) \\ \hline PaDiM & 96.11 $\pm$ 0.78 & 99.58 $\pm$ 0.25 & 93.14 $\pm$ 1.38 & 96.25 $\pm$ 0.72 & 99.22 $\pm$ 0.26 & 98.71 $\pm$ 0.47\\ \hline DFM & 90.89 & \textbf{99.96} & 84.61 & 91.64 & 98.63 & 97.38 \\ \hline WinCLIP & \textbf{98.08} & 97.98 & \textbf{99.75} & \textbf{98.86} & \textbf{99.32} & \textbf{99.83} \\\hline 
\end{tabular} \label{tab:MSTAR} \end{table}

\begin{table}[htb]
    \centering
    \caption{Average anomaly detection performance metrics on the image- and pixel-level for the HRSID dataset. All three models did not require training. DFM and WinCLIP are deterministic in their estimation of anomalistic data. PaDiM randomly selects 100 features from the backbone, and this model was used for three experimental trials, resulting in the standard deviation being shown for this model. The best average metric is bolded.}
    \renewcommand{\arraystretch}{1.2} 
    \begin{tabular}{|c|c|c|c|c|}
        \hline
        {Model} & {Pixel AUROC (\%)} & {Pixel F1 Score (\%)} & {Image AUROC (\%)} & {Image F1 Score (\%)} \\
        \hline
        {PaDiM} & \textbf{98.72} $\pm$ \textbf{0.36} & \textbf{27.96} $\pm$ \textbf{4.14} & \textbf{82.74} $\pm$ \textbf{5.59} & \textbf{98.88} $\pm$ \textbf{0.17} \\
        \hline
        DFM & 95.75 & 11.15 & 44.30 & 97.94 \\
        \hline
        WinCLIP & 75.75 & 4.04 & 65.52 & 98.19 \\
        \hline
    \end{tabular}
    \label{tab:HRSID}
\end{table}

\begin{figure}[htb]
    \centering
    \begin{subfigure}{0.22\textwidth}
        \centering
        \includegraphics[width=\linewidth]{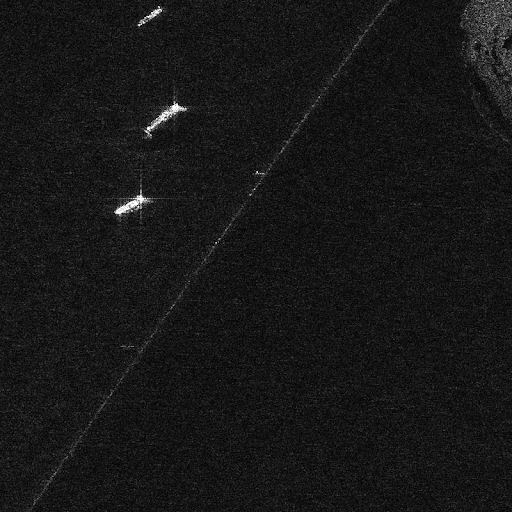}
        \caption{Input Image}
    \end{subfigure}
    \begin{subfigure}{0.22\textwidth}
        \centering
        \includegraphics[width=\linewidth]{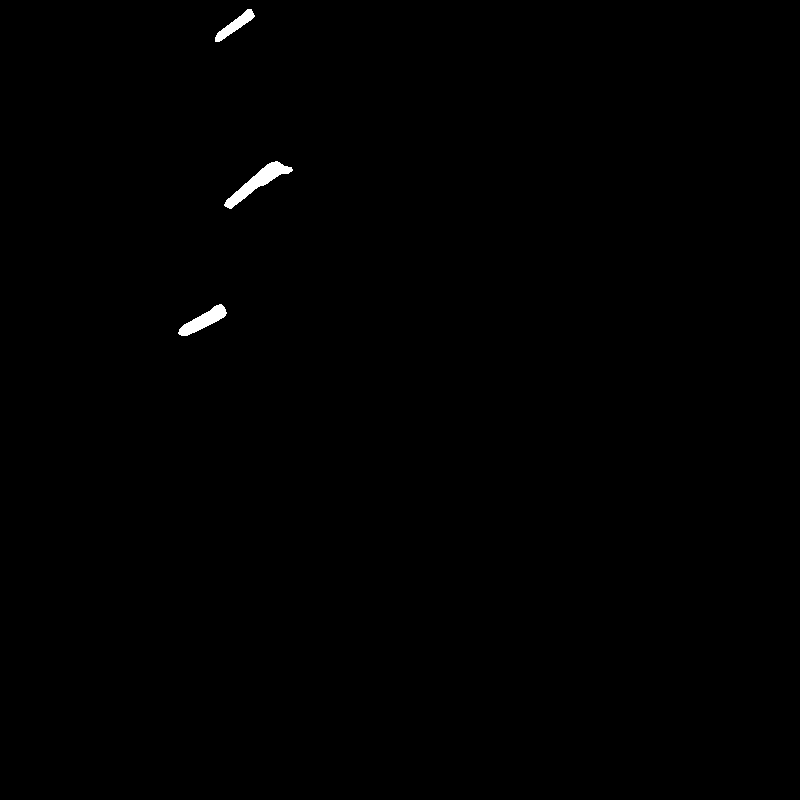}
        \caption{Ground Truth}
    \end{subfigure}
    \begin{subfigure}{0.22\textwidth}
        \centering
        \includegraphics[width=\linewidth]{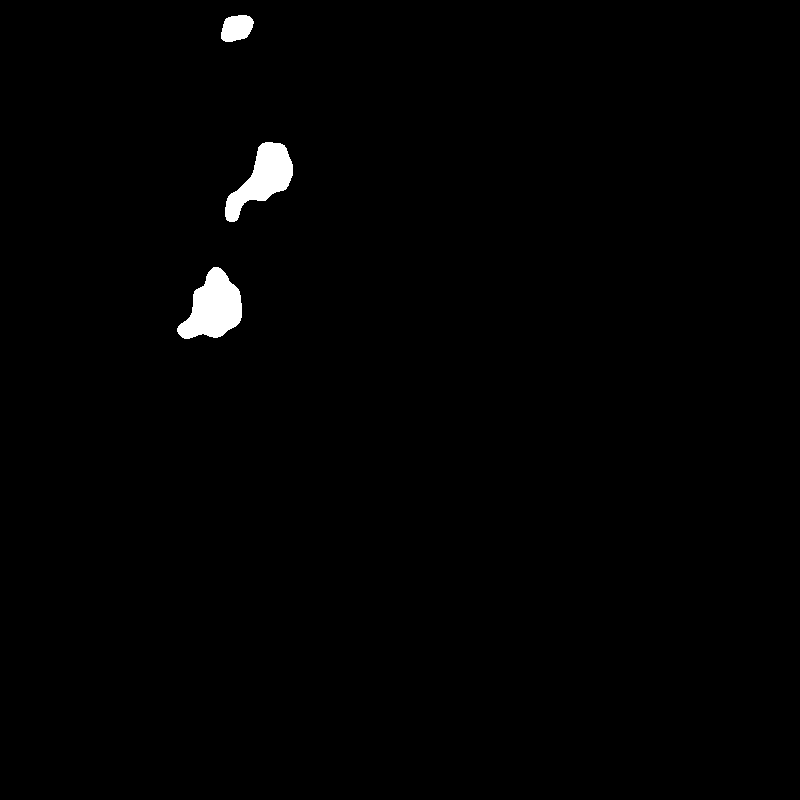}
         \caption{Predicted Mask}
    \end{subfigure}
    \begin{subfigure}{0.22\textwidth}
        \centering
        \includegraphics[width=\linewidth]{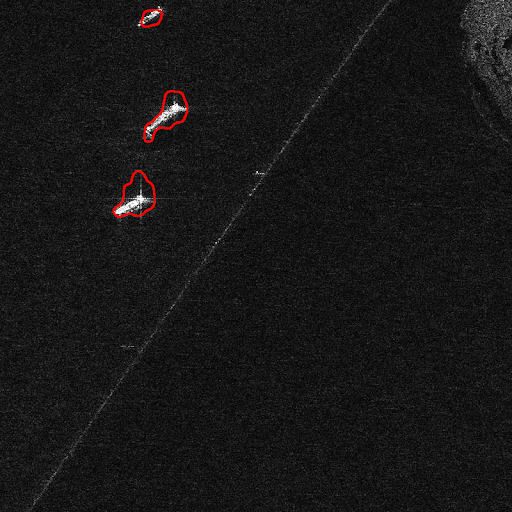}
         \caption{Detection Output}
    \end{subfigure}
        
    \caption{Example output resulting from applying PaDiM to a target chip in HRSID.}
    \label{fig:HRSIDMSTAR}
\end{figure}

For HRSID, metrics can be captured at the image and pixel level as shown in Table \ref{tab:HRSID}. We observe a similar trend that PaDiM outperforms the other models across the different metrics. WinCLIP and DFM are able to effectively detect if there is an anomaly present in the images, but the methods did not perform well for the pixel-level segmentation of the anomaly. More features will potentially provide a better representation of the normal samples to localize the anomalies in the images. All models do struggle with the pixel-level F1-score. HRSID dataset has images that contain small anomalies and mostly background, as shown in Figure \ref{fig:HRSIDMSTAR}. The AUROC may present an overly optimistic view for certain datasets with class imbalance \cite{saito2015precision,hancock2023evaluating}; therefore, the F1 score does provide insight to show that the models will need to be further improved to better localize the anomalies present in the images. 

\section{CONCLUSION}
\label{sec:conclusion}  
In this work, we presented SARIAD, a modular package for SAR imagery anomaly detection that builds on prior work from Anomalib \cite{akcay2022anomalib}. SARIAD is a tool that can leverage on SAR datasets for reproducible research by providing clear metrics and visualizations to compare pre-processing and models. Additionally, if no ``normal" images are present in the dataset, one can leverage the method introduced on the MSTAR to generate example images that can be leveraged to train the anomaly detection model. Future work includes several avenues: (1) incorporating additional SAR datasets, (2) refining the process of generating masks for ``normal" data, (3) adding advanced SAR anomaly detection models such as the SAR Automatic Target Recognition-X foundation model \cite{li2025saratr} and (4) integration of despeckling algorithms for more advanced pre-processing.

\acknowledgments 
 
This work was supported by the Laboratory Directed Research and Development program at Sandia National Laboratories, a multimission laboratory managed and operated by National Technology and Engineering Solutions of Sandia LLC, a wholly owned subsidiary of Honeywell International Inc. for the U.S. Department of Energy’s National Nuclear Security Administration under contract DE-NA0003525.

\bibliography{report} 
\bibliographystyle{spiebib} 

\end{document}